\documentclass{article}

\usepackage{arxiv}

\usepackage[utf8]{inputenc} 
\usepackage[T1]{fontenc}    
\usepackage{hyperref}       
\usepackage{url}            
\usepackage{booktabs}       
\usepackage{amsfonts}       
\usepackage{nicefrac}       
\usepackage{microtype}      
\usepackage{lipsum}
\usepackage{graphicx}
\usepackage{amsmath}
\usepackage[dvipsnames]{xcolor}
\usepackage{tipa}

\usepackage{comment}

\graphicspath{ {./images/} }

\title{Word Embeddings Are Capable of Capturing Rhythmic Similarity of Words}

\author{
 Hosein Rezaei \\
  Department of Electrical and Computer Engineering\\
  Isfahan University of Technology\\
  \texttt{hosein.rezaei@alumni.iut.ac.ir} \\
}
\begin{document}
\maketitle
\begin{abstract}
Word embedding systems such as Word2Vec and GloVe are well-known in deep learning approaches to NLP. This is largely due to their ability to capture semantic relationships between words. In this work we investigated their usefulness in capturing rhythmic similarity of words instead. The results show that vectors these embeddings assign to rhyming words are more similar to each other, compared to the other words. It is also revealed that GloVe performs relatively better than Word2Vec in this regard. We also proposed a first of its kind metric for quantifying rhythmic similarity of a pair of words. 
\end{abstract}

\keywords{Word Embeddings \and Rhythmic Similarity \and Word2Vec \and GloVe}

\section{Introduction}
Word embeddings (WE's) are one of the most popular outcomes of deep learning algorithms in the realm of natural language processing. They are applied in a broad range of downstream applications such as question answering, word sense disambiguation, reading comprehension, summarization, etc. This is largely due to their success in capturing semantics of words in terms of finite and relatively low dimensional vectors. In a simple way, they are able to capture meaning of words without feeding any external knowledge, just by going through a raw corpus of words. In fact they do keep track of each word's context.

In addition to this influential characteristic, WE's seem to have yet another, almost neglected characteristic of capturing rhythmic similarity between words, especially in the literature corpora. But what does that mean? To answer this question, it is needed to first define Rhythmic Similarity and then determine what's the meaning of capturing. 

\subsection{What is Rhythmic Similarity (RS)?}
For the purpose of this project, we define two given words as rhythmically similar, if they are said (pronounced) with similar sounds, specially at their ending or beginning. For example, the words 'doom' and 'gloom' have rhythmic similarity because they both are ended with the sound: /\textipa{u:m}/, another example is 'cheese' and 'peas' \footnote{The example is taken from BBC website: \href{https://www.bbc.co.uk/bitesize/topics/zjhhvcw/articles/zqjgrdm}{https://www.bbc.co.uk/bitesize/topics/zjhhvcw/articles/zqjgrdm}}. Even words like 'tend' and 'tell' are rhyming words, to a lesser degree though. Thus we will need a metric to assess this degree, and this will be introduced in section \ref{sec:metrics}

Having defined RS in the above terms, this research tries to investigate if WE's can capture RS? Here by capturing we mean, if the geometric relationship between vectors do represent any other relation as well? In particular, if the proximity of vectors represents the RS as well? and if yes, to what extent? This is analogous to the famous concept of capturing semantic similarities, whereby we have witnessed that vectors embedded in a neighboring space tend to have similar meanings. Considering the above question as the topic of this research, the next step is to describe how we want to seek an answer for it. Thus we define a short hypothesis to make our research process more clear.

\subsection{Hypothesis}
\label{sec:hypothesis}
The hypothesis we are going to investigate in this study is that "Word embeddings assign similar vectors to the words which are rhythmically similar". We want to see if this is true and if yes, to what extent? We investigate this on several datasets and texts including literature and non-literature ones. We also examine this from the perspective of several WE's to see how different they act on this regard. 

In the remaining of this paper, a review on the literature is presented in the section \ref{sec:literature}, the methodology is explained in section \ref{sec:methodology}, the experiments are introduced in section \ref{sec:experiments}, section \ref{sec:results} demonstrates their results, and finally section \ref{sec:conclude} concludes the discussion and presents some future works.

\section{Literature Review}
\label{sec:literature}
Digital representation of natural language (mostly words and sentences) for computer processing has been a major challenge from the early days of research in this area\cite{camacho2018word}. Simple intuitive solutions such as one-hot encoding had left researchers with sparse vectors of long length, at the order of millions, which were challenging to process. Used as features, such trivial vectors were representing nothing but the presence of words. Then, semantic space models, LDA, LSI, and LSA emerged and led to the use of more compact vectors \cite{blei2003latent}, \cite{dumais1994latent}, \cite{landauer1998introduction}. Later on, the introduction of Neural Networks triggered the development of hugely successful models such as Word2vec \cite{Mikolov2013} and GloVe \cite{pennington2014glove}. And more recently, with the boom of Deep Learning, many more embeddings are brought about, among which BERT\cite{devlin2018bert}, ELMO\cite{Peters2018}, and GPT\cite{radford2019language} have gained a huge reputation.

Getting inspiration from the primitive models and making revisions of WE’s for different purposes, has been a major trend in recent years. For example, Jameel and Schockaert in \cite{jameel2016d} presented D-GloVe which represents each words with a probabilistic density to reflect the uncertainty of meaning based on both the frequency of words in the corpus and their informativeness. In another work\cite{shi2017jointly}, Bei Shi et.al proposed STE model which learns WE’s and topic models at the same time. Their architecture utilizes a generating function to represent topics with vectors, and also improves the traditional skip-gram model of Word2vec by EM-negative sampling. They demonstrated that such a model is able to generate more meaningful WE’s which also represent coherent topics.  

However, all of the above mentioned systems are centered around semantics. Consequently, the phrases "word similarity" and "similar words" are extensively used in the literature. In this work, we aim to add a finer level of precision, and draw attentions to the fact that the notion of similarity can be defined from different perspectives. Although the above terms are used normally for referring to "word semantic similarity" and "semantically similar words" respectively, there are other aspects such as "morphological" or "rhythmic" similarity as well. And these other aspects are important to be studied. 
In regard to the importance of various types of similarity, it would be useful to take a look at one of the groundbreaking papers of the field. In the very famous paper of Mikolov et.al \cite{Mikolov2013} by which Word2vec was introduced to the scientific community, they have written: 

\begin{quote}
\textit{
    "We use recently proposed techniques..., with the expectation that not only will similar words tend to be close to each other, but that words can have \textbf{multiple degrees of similarity}"
}
\end{quote}. 
And then, to exemplify another degree of similarity (in addition to semantics), Mikolov mentions the "similarity of endings": 

\begin{quote}
\textit{
    "This has been observed earlier in the context of inflectional languages - for example, nouns can have multiple word endings, and if we search for similar words in a subspace of the original vector space, it is possible to find words that have similar endings". 
}
\end{quote}

Although, this paper has been one of the top referenced and most influential studies in the literature, this particular statement has been rarely noticed. To the best of our knowledge, no further expansion on this direction has been done, neither by the authors of that paper nor by any other researchers. Even earlier observations cited in the above statement, didn't try to quantify this claim and assess its statistical significance. Thus, we think this is a valuable and novel contribution to pursue.

\section{Methodology}
\label{sec:methodology}
In this section we're going to elucidate in details how we will try to verify our hypothesis (section \ref{sec:hypothesis}). First, we formulate our method and then introduce RS metrics as it is the the core concept of this work. 

In order to verify our hypothesis for a specific WE on a specific corpus C, we first generate WE vectors for C. Then using a clustering method, we put the nearer vectors and corresponding words into $k$ clusters. Then we propose some metrics to quantify the degree of Rhythmic Similarity (RS). Using that metric, we measure RS of each pair of words inside a cluster $c_i$, $0<i\leq k$, and then calculate the average RS of each cluster, $\overline{RS_{c_i}}$, $0<i\leq k$. 
\begin{equation}
\label{eq-cluster-rs}
    \overline{RS_{c_i}} = \frac{\sum_{x=1}^{n_i}\sum_{y=x+1}^{n_i}RS(w_{i,x}, w_{i,y})}{\frac{n_i*(n_i-1)}{2}}
\end{equation}
where $n_i$ is number of words inside $i$th cluster, $w_{i,j}$ is the $j$th word of the $i$th cluster, and $RS(w_i, w_j)$ is a real number between 0 to one representing how similar the words $w_1$ and $w_2$ are to each other from rhythm perspective. See section \ref{sec:metrics} for a formal definition.

Then we can take an average over all clusters, to have an overall sense of how homogeneous the clusters are in terms of RS. 

\begin{equation}
    \label{eq-curpus-rs}
    \overline{RS} = \frac{\sum_{i=1}^{k}\overline{RS_{c_i}}}{k}
\end{equation}
where $k$ is total number of clusters.
If this number tends towards one, this means that the majority of clusters have a high degree of RS, which in turn means the majority of words inside each cluster have high RS. However, since this average is taken over many numbers ($k$ is in the order of thousands or more), many numbers might neutralise each other, and we lose information. Thus, we can divide the range from 0 to 1 into some bins, and count the number of clusters which fall in the same bin in terms of $RS_{c_i}$. Drawing the histogram, if bins near 1 are occupied by more clusters, this proves that WE has been successful in assigning vectors near each other in the embedding spaces to the rhymes and assonances. 

In order to assess the extent of this behaviour better, a baseline is needed. Therefore, we randomly put words inside $k$ new clusters, and compute the average RS of random clusters $\overline{RS_{rc_i}}$ , $0<i\leq k$. In other words, we compute the equation (\ref{eq-cluster-rs}) again, but this time over the random clusters instead of clusters based on WE. Then, comparing the $\overline{RS_{c_i}}$ with the baseline, if they tend to have larger values, this means that WE clusters have higher RS compared to random clusters and this proves our hypothesis. If we found that a majority of WE clusters have the same or lower RS as that of random clusters, this means that WE vectors aren't able to capture RS significantly.

Many parameters of the above experiment can be altered to investigate their effect on the hypothesis. For example we can use different corpora, different clustering methods, different WE's with different parameters, and so on.
But more important than any other aspect, a metric for measuring RS can affect our evaluation. Thus it is needed to define some metrics first. 

\subsection{Metrics for Rhythmic Similarity}
\label{sec:metrics}
In the early observations of this project, we witnessed that words whose vectors have fallen within the same cluster, are almost similar in their ending or beginning letters. But how similar they are? And how we can compare two clusters in this regard? In order to answer this question we devised a simple formula that acts on the basis of corresponding letters in the input pair. Specifically, it takes number of common letters in the same positions into account, starting from the end of words. This is then divided by total number of letters. 

\begin{equation}
\label{eq-rs-basic}
RS _{}(w_1,w_2)= {\frac 
    {
    \sum _{i=1}^{cl}
    \begin{cases} \mbox{1,} & \mbox{if } letter(w_1, i)==letter(w_2, i) \\ \mbox{0,} & \mbox{otherwise} \end{cases}
    }
    {max(length(w_1), length(w_2)) }
}
\end{equation}

where $cl$ is the common length of $w_1$ and $w_2$ i.e. $min(length(w_1) , length(w_2))$ and $letter(w, i)$ is the $i$th letter of $w$ starting from the end. So for example, $RS(h\textcolor{teal}{ustle}, b\textcolor{teal}{ustle})=5/6=0.83$, $RS(h\textcolor{teal}{o}l\textcolor{teal}{y}, technol\textcolor{teal}{o}g\textcolor{teal}{y})=2/10=0.2$ $RS(rotation, positions)=0/9=0$ because starting from the end, they have no letter in common. The range of values this metric produces is from 0 (entirely dissimilar) to 1 (totally similar). These values can be multiplied by 100 and expressed in terms of percentage.

Clearly, this metric is naive and not very precise, since it is too sensitive to the location of letters. As we saw in the last example, a single s letter in the end of the word, made all other letters misaligned and set RS to zero, whilst they are highly similar, intuitively. This metric also does not distinguish between consonants and vowels, whereas we know that vowels play a more significant role in rhythm compared to consonants. Consider for example, $RS(r\textcolor{teal}{e}f\textcolor{teal}{i}n\textcolor{teal}{e}, d\textcolor{teal}{e}v\textcolor{teal}{i}s\textcolor{teal}{e})=3/6=0.5$ which by intuition seems relatively low, but if we consider only vowels, the new $RS$ would be $3/3=1$ which seems more realistic. 
Syllabification is also a very influential parameter in measuring assonance between words, but is not considered in this primitive metric.

Another idea is to compare IPA transcription of the words since they represent phonetic similarities better. For example, for words "cheese" and "peas", $RS(che\textcolor{teal}{e}se, p\textcolor{teal}{e}as)=1/6=0.17$ but if we define:

\begin{equation}
\label{eq-rs-ipa}
RS_{IPA}(w_1, w_2)=RS(IPA(w_1), IPA(w_2))
\end{equation}

where $IPA(w)$ means IPA representation of the word $w$, then we have 
$RS_{IPA}(cheese , peas) = RS($\textipa{tS\textcolor{teal}{i:z}}
$,$
\textipa{p\textcolor{teal}{i:z}}$)=3/5=0.6$
and this value seems more reasonable.

However, for the purpose of this study, the basic metric (\ref{eq-rs-basic}) is adequate and preferable in terms of performance in run time.

\section{Experiments}
\label{sec:experiments}
In this section we describe what experiments are designed to verify hypothesis practically. 
\subsection{Corpora}
The early observations of this project were based on Quran corpus \cite{zeroual2016new}, whereby we witnessed that words whose vectors have fallen within the same cluster, are almost similar in their ending or beginning letters\cite{linkedin}.  

However, we evaluated our experiments on two other corpora as well to see if the same phenomenon can be witnessed in other texts or not. The second corpus is taken from a collection of poetic texts in English, from Gutenberg dataset \cite{bird2009natural}. The third corpus is taken from CNN news \cite{lins2019cnn}, which we know contains less rhythmic content. Since Quran contains around 17K unique words, we taken the size of the other datasets in a way to cover, roughly, the same number of unique words, so to keep our comparison balanced and unbiased. See table (\ref{tab:corpora} for more details.

\begin{table}
 \caption{Overview of corpora}
  \centering
  \begin{tabular}{llll}
    \cmidrule(r){1-4}
    Name     & \#All words     & \#Unique words & Average Word Length \\
    \midrule
    Quran & 82,624  & 17,627 & 9.77     \\
    Gutenberg     & 166,000  & 16,545 & 6.63      \\
    CNN     & 529,359 & 26,279 & 7.45  \\
    \bottomrule
  \end{tabular}
  \label{tab:corpora}
\end{table}

\subsection{Word Embeddings}
In this study we examined Word2Vec \cite{Mikolov2013} and GloVe \cite{pennington2014glove} to see how they capture RS between words. Nevertheless the same methodology can be applied to other WE's as well.

For generating Word2Vec vectors, we used the Gensim library in Python and for GloVe, the original implementation by \cite{pennington2014glove}. For both methods we set the size of vectors=100, size of window is set to 15, and min\_count=1 which means all words are taken into account. Other parameters are set to default as defined in version 4.1.2 of Gensim and latest version of GloVe\footnote{see \href{https://github.com/stanfordnlp/GloVe/blob/d470806c6c23c6698a02d533c0497f4c25c98a77/demo.sh}{https://github.com/stanfordnlp/GloVe/blob/d470806c6c23c6698a02d533c0497f4c25c98a77/demo.sh}}.

\subsection{Clustering}
For clustering, we used K-means with k = 1000 and cosine distance as the measure for the geometric distance of vectors in the embedding space. However, other clustering methods, other values for k, and other distance measures can be investigated in future works. 

\section{Results}
\label{sec:results}
The results of the experiments can be demonstrated in two forms. The simplest way is to compare $\overline{RS}$ of clusters when made by WE's with the same metric when clusters are made randomly. This is shown in table \ref{tab:result}. As you can see, the $\overline{RS}$ of random clusters is approximately 5$\sim$6, meaning that on average, words inside each cluster have been 5 to 6\% similar to each other in terms of rhythm and assonance. Considering these numbers as the baseline, it is obvious that clusters of Word2Vec and GloVe have been more homogeneous rhythmically, as their $\overline{RS}$ have been higher than baseline. If we compare corpus-wise, we expect the Gutenberg clusters to bear more $\overline{RS}$ because they are more poetic, and this is the case for both random clusters and GloVe. For Word2vec, however, this isn't true which can be attributed to the global nature of Word2Vec and proximity of rhymes in poems. In other words, in poetic texts, rhymes occur near each other, but Word2Vec mixes all words of the corpus and doesn't care about the relative position of words. In contrast, GloVe, using a local context window moving over words, makes it possible for the $RS$ of neighboring words to be reflected in the vectors. This trend is more obvious in Quran, as GloVe clusters have gained $\overline{RS}=12$, as twice as Word2Vec and random clusters. This can be also due to the fact that Quran contains more rhythmic content, as almost all adjacent sentences end with rhymes. An evidence for this is that even $\overline{RS}$ of random clusters of Quran is 6.90, visibly higher than that of CNN and Gutenberg, which are 5.48 and 5.80, respectively.

\begin{table}
 \caption{Average RS ($\overline{RS}$) of clusters}
  \centering
  \begin{tabular}{llll}
    \cmidrule(r){1-4}
    Name     & Quran     & CNN & Gutenberg \\
    \midrule
    Word2Vec & 7.26  & 6.73 & 6.03     \\
    GloVe     & 12.16  & 7.03 &  7.20      \\
    Random     & 6.90 & 5.48 & 5.80  \\
    \bottomrule
  \end{tabular}
  \label{tab:result}
\end{table}

\begin{figure}
  \centering
  \includegraphics[width=0.6\textwidth]{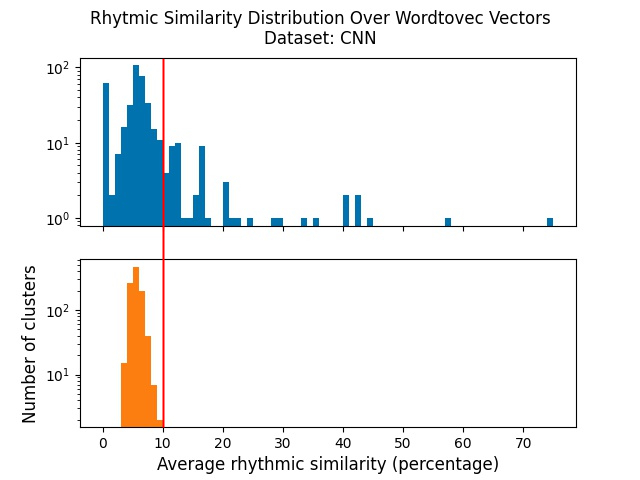}
  \caption{The bottom chart in orange shows that all random clusters have had $\overline{RS}$ below 10\%. In contrast, the blue chart shows that Word2Vec clusters are more distributed over higher ranges of RS. The chart is log-scaled.}
  \label{fig:cnn-w2v}
\end{figure}

\begin{figure}
  \centering
  \includegraphics[width=0.6\textwidth]{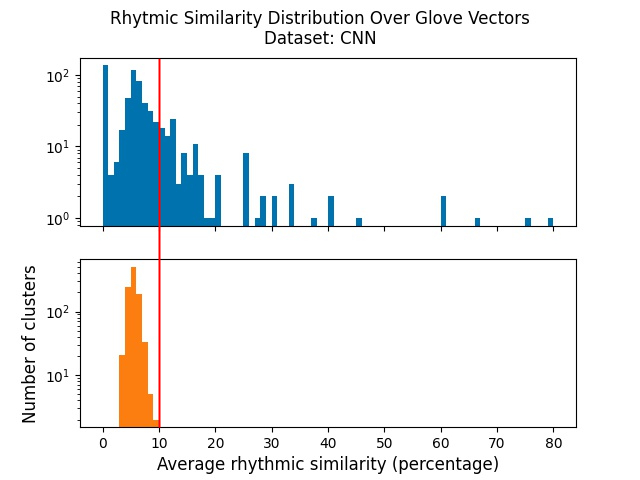}
  \caption{The blue chart is more spread over x axis compared with the orange one and also compared with the blue chart of figure \ref{fig:cnn-w2v}. This suggests that GloVe clusters have more RS than random or Word2Vec clusters.}
  \label{fig:cnn-glove}
\end{figure}

\begin{figure}
  \centering
  \includegraphics[width=0.6\textwidth]{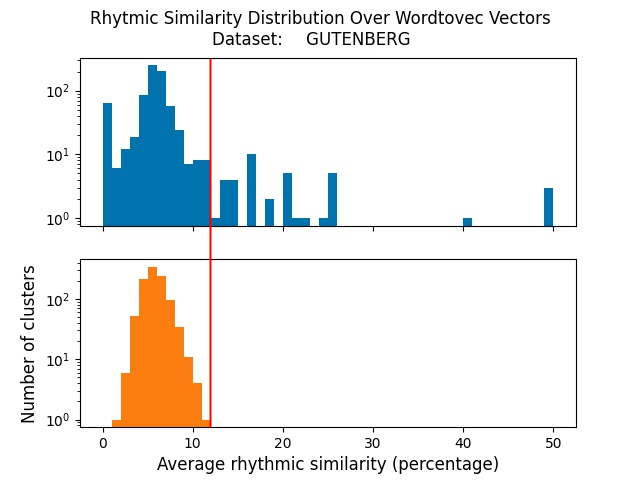}
  \caption{Results for the Gutenberg corpus. RS is more distributed for Word2Vec clusters (the blue chart), compared with random clusters (the orange chart).}
  \label{fig:poems-w2v}
\end{figure}

\begin{figure}
  \centering
  \includegraphics[width=0.6\textwidth]{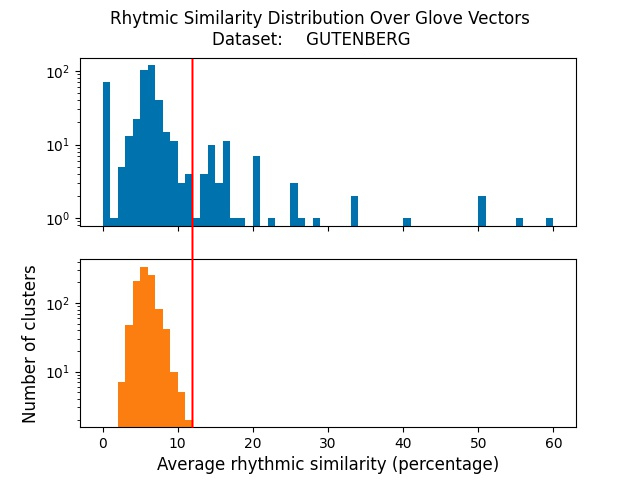}
  \caption{Outcome of Gutenberg corpus for GloVe. Again, compared with random clusters and compared with Word2Vec in figure \ref{fig:poems-w2v}, this chart demonstrates that GloVe performs better in capturing RS.}
  \label{fig:poems-glove}
\end{figure}

\begin{figure}
  \centering
  \includegraphics[width=0.6\textwidth]{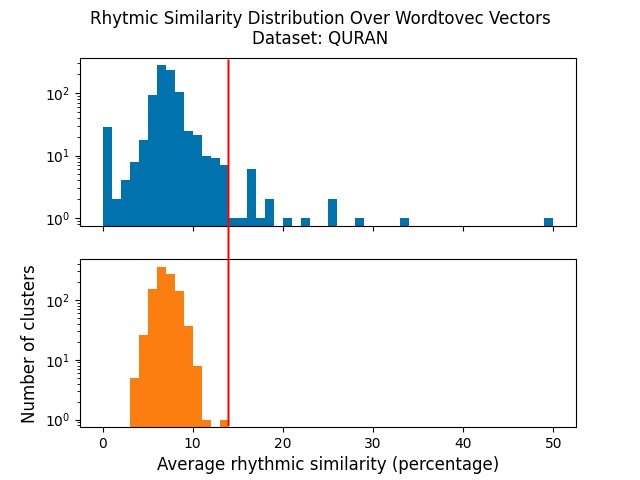}
  \caption{Output of experiment for Quran corpus. Although the spread is not as much as previous corpora (figures \ref{fig:cnn-w2v} and \ref{fig:poems-w2v}), but the area below the blue chart is clearly larger than that of random chart. }
  \label{fig:quran-w2v}
\end{figure}

\begin{figure}
  \centering
  \includegraphics[width=0.6\textwidth]{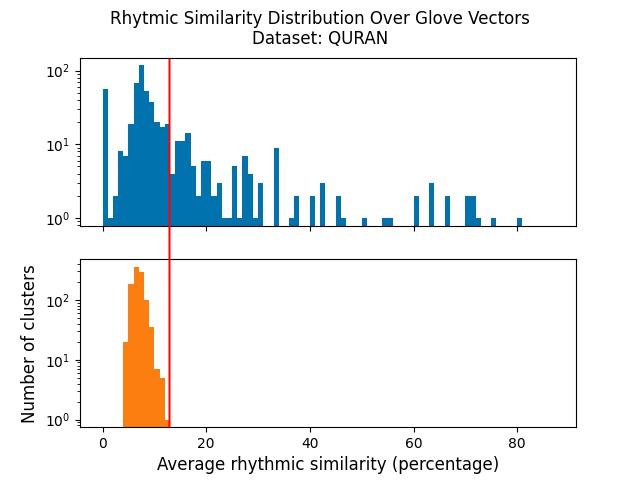}
  \caption{Output of experiment for the Quran corpus and the GloVe clusters. The same trend can be seen, as in the other two corpora.}
  \label{fig:quran-glove}
\end{figure}

The above method, i.e. computing average, has a drawback that even if some clusters have high $\overline{RS}$, their effect on average might be neutralised by other clusters having low $\overline{RS}$. Therefore, another way to demonstrate the results, is to count how many WE clusters have $\overline{RS}$=100\% and compare this value with that of random clusters. Repeating this over all possible values of RS ($0\le{RS}\le 100$) gives us a sense of how well that WE is performed in assigning nearer vectors to rhymes. Such data can be best represented with histograms. For drawing them, we broken the range from zero to 100 into 100 bins, and counted the number of clusters whose $\overline{RS}$ falls inside the range of each bin. This stats is computed for both random clustering and WE clustering. Regarding CNN corpus, the figures \ref{fig:cnn-w2v}
and \ref{fig:cnn-glove} show this stats for Word2Vec and GloVe respectively. For Gutenberg corpus, figures \ref{fig:poems-w2v} and 
\ref{fig:poems-glove} and for Quran, figures \ref{fig:quran-w2v} and \ref{fig:quran-glove} show the similar stats, respectively. 

The same patterns as we saw in mean RS, can be discerned here. In all charts, the random clusters (in orange color) hardly spread beyond 15. This means that when we cluster words randomly, the $\overline{RS}$ of clusters is at most 15\% (the red line). Contrastingly, when clustered by Word2Vec or GloVe (the blue bars), the $\overline{RS}$ spreads over a wider range and the area below the bars is noticeably larger. This means that WE clusters are more homogeneous in terms of rhymes. 

And overall, we can say that WE's tend to assign similar vectors to the rhyming words and thus caused them to be placed in the same clusters. 

\section{Conclusion}
\label{sec:conclude}
In this study, we tried to figure out if WE's are capable of capturing RS and how much? We discussed that this property of WE's is mentioned in the pioneering works of the field but never got enough attention. We formulated a hypothesis and investigated it on several corpora and several WE's. The results showed that when we cluster words based on their WE vectors, word having higher RS tend to be in the same clusters. Even though the words in a cluster aren’t absolute rhymes, but the probability of a rhyme pair to be in a WE cluster is more than that of random clusters. As future works, one can examine more WE's and verify the same hypothesis on them. Also more fine-grained RS metrics can be proposed, specifically to take syllabification into account. Another interesting line of research is to develop new WE's that intentionally capture RS, i.e. their objective function would be maximizing RS between words. 
The code of this research is publicly available at Github\footnote{see \href{https://github.com/HRezaei/rhymes}{https://github.com/HRezaei/rhymes}}.

\newpage

\bibliographystyle{unsrt}  
\bibliography{references}  


\end{document}